\RequirePackage{fix-cm}
\documentclass[smallextended]{svjour3}       
\smartqed  
\usepackage{graphicx}
\usepackage{booktabs} 
\usepackage[ruled]{algorithm2e} 
\usepackage{amsmath}
\usepackage{cases}
\usepackage{bbm}
\usepackage{caption}
\usepackage{enumitem}
\usepackage{cite}
\usepackage[numbers]{natbib}
\usepackage{CJKutf8} 
\usepackage{tablefootnote}
\usepackage{enumerate}
\usepackage{multirow}
\usepackage{hyperref}
 
\usepackage{float} 
\usepackage{subfigure}
\usepackage{setspace}
\usepackage{soul,color}
\sethlcolor{yellow}
\soulregister\cite7 
\soulregister\citep7 
\soulregister\citet7 
\soulregister\ref7 
\soulregister\pageref7 
\usepackage[misc]{ifsym}

\usepackage{bbding}
\usepackage{amsmath,amssymb,amsfonts}

\begin{document}

\title{Prediction, Selection, and Generation:
A Pre-training Based Knowledge-driven Conversation System
}
\subtitle{}


\author{Cheng Luo\textsuperscript{1} \and
	Dayiheng Liu\textsuperscript{2} \and
	Chanjuan Li\textsuperscript{2} \and
	Li Lu\textsuperscript{2} \and
	Jiancheng Lv\textsuperscript{1} 
}




\institute{ 
\Letter  Jiancheng Lv\\
\hspace*{1em} \email{lvjiancheng@scu.edu.cn}\\
\Letter  Li Lu\\
\hspace*{1em} \email{luli@scu.edu.cn}  
\at
\textsuperscript{1} College of Computer Science, State Key Laboratory of Hydraulics and Mountain River Engineering, Sichuan University, 610065, Chengdu, China\\
\textsuperscript{2} College of Computer Science, Sichuan University, 610065, Chengdu, China
}

\date{Received: date / Accepted: date}

\maketitle

\begin{abstract}
In conversational systems, we can use external knowledge to generate more diverse sentences and make these sentences contain real knowledge.
Traditional knowledge-driven dialogue systems include background knowledge by default when generating responses, which has a gap with reality conversations.
At the beginning of a reality conversation, the system does not know what background knowledge is needed, which requires the system to be able to select the appropriate knowledge from the knowledge graph or knowledge base for generation based on the context.
In this paper, we combine the knowledge graph and pre-training model to propose a knowledge-driven conversation system and a dialogue generation model Bert2Transformer.
The system includes modules such as dialogue topic prediction, knowledge matching, and dialogue generation.
The Bert2Transformer model does a great job of using context and knowledge to generate fluent responses.
Based on this system, we study the performance factors that maybe affect the generation of knowledge-driven dialogue: topic coarse recall algorithm, number of knowledge selection, generation model choices, etc., and finally make the system reach state-of-the-art.
\end{abstract}

\section{Introduction}

\begin{figure*}[h]
	\centering
	\centerline{
	    \includegraphics[width=1\linewidth]{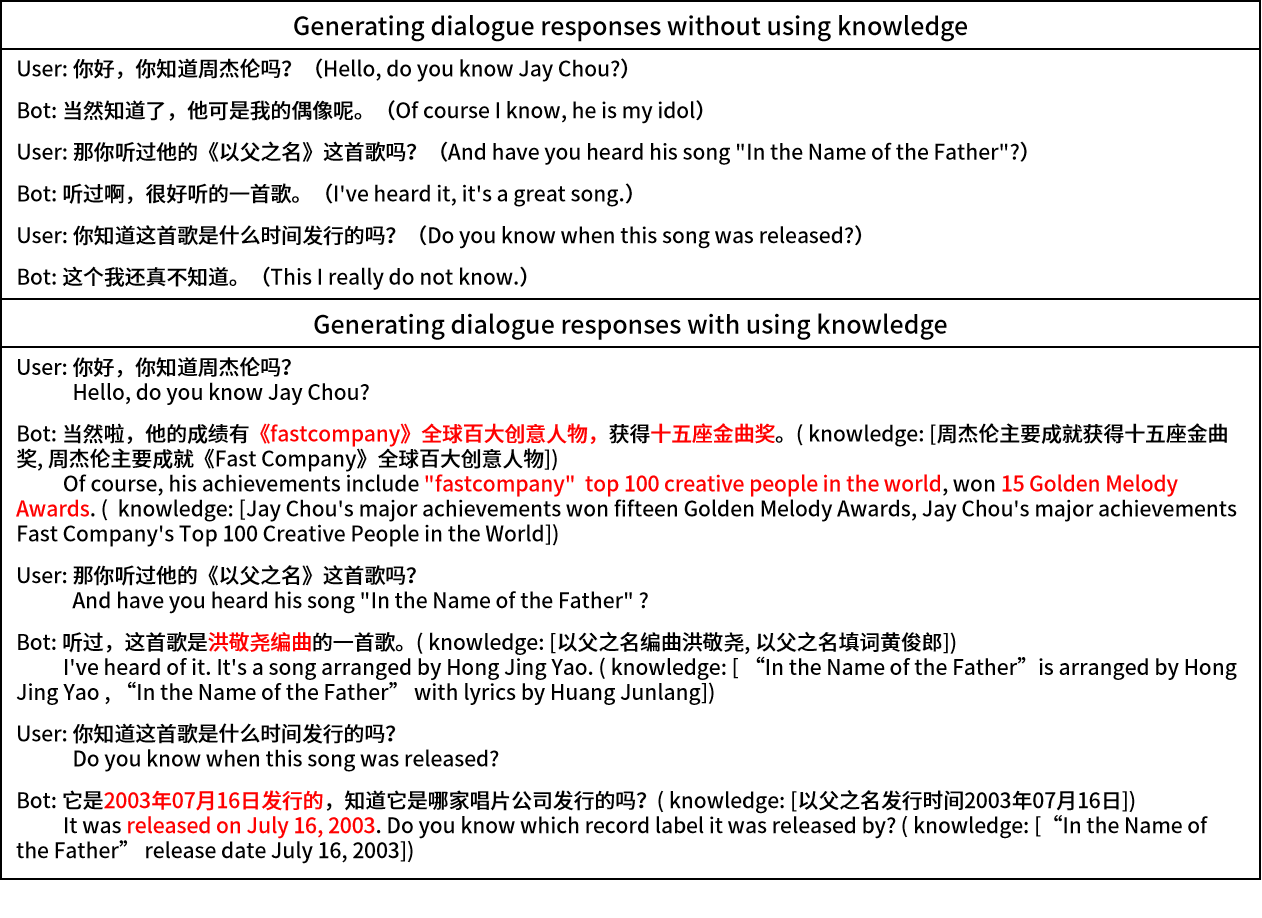}}
    \caption{Two examples of using knowledge-driven or not. 
    Both examples use the same model that is trained in our experiments, the first example is generated without adding external knowledge and the second example is generated with adding external knowledge.
    Sentences marked in red indicate the result of knowledge-driven generation.}
	\label{fig:Knowledge to show}
\end{figure*}

It has been a long-term goal of artificial intelligence to deliver human-like conversations, where background knowledge plays a crucial role in the success of conversational systems \cite{DBLP:journals/corr/abs-2011-11928, 10.1145/3397271.3401419, 10.1145/3394486.3403258}, as shown in Figure \ref{fig:Knowledge to show}.
The arrival of the pre-training model era has also greatly promoted the generation of dialogue in the open-domain.
For example, CDial-GPT2 \cite{DBLP:conf/nlpcc/WangKZHJZH20} and OpenAI GPT-2 \cite{radford2019language} have demonstrated that transformer trained on a large-scale dataset can capture long-term dependencies in textual data and generate fluent diverse text.
Such models have the capacity to capture textual data with fine granularity and produce output with a high-resolution that closely emulates real-world text written by humans.
If we want these models to generate texts that contain correct realistic knowledge or controllable content, we can incorporate additional knowledge information to generate.
\par
Integrating knowledge into dialogue generation can make the generated results more diverse and controllable\cite{DBLP:conf/emnlp/LiuGYFSJLD20}.
For example, incorporating laughter knowledge into a Generative Model can generate humorous dialogue \cite{DBLP:conf/acl/ZhangLLL20}.
In an open-domain conversation system, it is very important but challenging to use background knowledge for effective interaction.
Background knowledge can be expressed as a knowledge graph\cite{zhang2021generalized}, unstructured text \cite{DBLP:conf/aaai/GhazvininejadBC18} or a descriptive corpus\cite{wang2015constructing}.
\par
Previous knowledge-driven conversation systems usually assume that background knowledge is already known when generating responses.
In contrast, knowledge-driven conversation systems in real application scenarios need to select appropriate knowledge based on the dialogue context and generate fluent responses, just like human chats.
In this paper, we propose a knowledge-driven conversation system with knowledge selection capability. 
To explore the factors that affect the generation effect of this system, we use different models to conduct experiments on the dataset KdConv \cite{DBLP:conf/acl/ZhouZHHZ20} under this system, and finally find the main factors affecting the generation effect, and make the system has reached state-of-the-art results.
\par
Our contributions are threefold:
(1) We propose a general knowledge-driven dialogue generation system which covers the entire process from topic prediction, knowledge selection and dialogue generation.
(2) We study the relevant factors affecting the correct knowledge selection rate and find that the highest accuracy is achieved by using the LAC algorithm for rough recall and Sentence-Bert \cite{DBLP:conf/nlpcc/WangKZHJZH20} is better than Pairwise Model for sorting for fine recall.
(3) We investigate the factors affecting the generation effect and propose the Bert2Transformer model. After comparing some factors such as generation model selection, multi-task, and knowledge number selection, we find that using the Bert2Transformer model works better than CDial-GPT2, and in this case, selecting 3 pieces of knowledge generates better results than selecting 1 piece of knowledge.

\section{Related Work}
\textbf{Open-domain Conversations Generation.}
In traditional open-domain conversations, the model predicts the next sentence given the previous sentence or sentences in the conversation \cite{vinyals2015neural, lei2018sequicity, jin2018explicit}.
The responses generated by such systems tend to be safe response \cite{DBLP:conf/naacl/LiGBGD16} and do not contain knowledge that does not appear in the context.
In order to generate multiple diverse responses, many approaches resort to enhanced beam search \cite{DBLP:conf/naacl/LiGBGD16, li2016simple}.
And some studies allow the model to generate a diversity of responses rather than in the post-processing stage.
For example, some researchers augment the Seq2Seq model with a multi-mapping mechanism to learn the one-to-many relationship for multiple diverse response generation \cite{DBLP:conf/ijcai/ChenPWXW19, DBLP:conf/emnlp/QiYGLDCZ020}.
However, these efforts do not allow the model to generate responses that contain knowledge not mentioned in the context.

\noindent
\textbf{Application of Knowledge.}
Knowledge is now used in a variety of tasks.
Some researchers use knowledge to pre-train a model (ERNIE) so that it can use the knowledge information for downstream tasks \cite{DBLP:conf/acl/ZhangHLJSL19}.
And some researchers enrich the state-of-the-art neural natural language inference models with external knowledge \cite{DBLP:conf/acl/InkpenZLCW18, liang2020pirhdy}.
Other researchers introduce a neural reading comprehension model that integrates external knowledge, encoded as a key value memory \cite{DBLP:conf/acl/FrankM18}.
Incorporating external knowledge can give more information to the model, which is believed to be the trend.

\noindent
\textbf{Knowledge-driven Conversation.}
On dialogue tasks, there are also a number of studies that focus on making models to incorporate knowledge to produce more diverse and knowledge-inclusive responses.
Some of them focus on the choice of knowledge.
For examples, DiffKS \cite{DBLP:conf/emnlp/ZhengCJH20} propose a difference-aware knowledge selection method to facilitate the selection of more appropriate knowledge.
And some studies focus on conversation generation.
MGCG \cite{DBLP:conf/acl/LiuWNWCL20} is an architecture of multi-goal driven conversation generation framework that requires the input of goals, knowledge, and other information to generate conversation responses.
This framework requires users to set chat knowledge and goals to generate responses.
In actual dialogue, however, we need a model or system to select appropriate knowledge based on background and history to generate dialogue by itself.
Some works focus on this issue such as \citep{DBLP:conf/aaai/GhazvininejadBC18}, in which knowledge selection is considered and the model performs simple knowledge selection before generation, but the selection method is very crude and each generation selects a large number of different types of knowledge for generating a response, thus introducing a large amount of noise and reducing the generation quality.
\par In this paper, we propose a knowledge-driven conversation system that covers precise knowledge selection and generates fluent responses that contain knowledge content.
Based on this system, we explore the key factors affecting its performance.

\section{Methodology}
In this section, we introduce our experimental models and methods. 
We experiment with multiple elements, which intuitively work for dialogue generation, to explore how each element is critical to the task.

\subsection{Task Formulation}
In the case of our task, there are differences between the training and inference phases.
In the training phase, we have three models to train: a topic prediction model, a knowledge matching model, and a conversation generation model.
We process the data into the standard training data needed for each model, so there is no connection between these models.
But the inference phase is more like our conversations in reality, where we only have the historical conversational corpus, which requires the individual models to work together to generate responses with real knowledge.
So it is different from traditional multi-turn dialogue, knowledge-driven multi-turn dialogue should include knowledge selection work in addition to dialogue generation.
\par
We suppose that we have datasets $D_{kc}=\{(s_i)\}^N_{i=1}$ and $K_{kc}=\{k_j\}$ from KdConv, where $s_i= \{u_{i,1} ,...,u_{i,n}\}$ represents a dialogue scene with $n$ rounds utterances, $N$ denotes the number of dialogue samples, $k_j$ stands for relevant knowledge base or knowledge graph.
Our goals are to use these data to train the system to generate reasonable responses and explore the factors that affect its performance.

\subsubsection{Training Phase}
In the training phase, we reconstruct the datasets $D_{kc}$ and $K_{kc}$ as different training datasets $D_{topic}$, $D_{kg}$, $D_{cg}$ for different models.

\noindent
\textbf{Topic Prediction Model.}
$D_{topic}=\{(h_i,m_i)\}^N_{i=1}$, where $N$ means there are $N$ pieces of training data, $h_i= \{u_{i,1} ,...,u_{i,n}\}$ represents a conversation context with $u_{i,n}$ as a utterance, and $m_i$ is the label of the topic from the dataset $K_{kc}$.

\noindent
\textbf{Knowledge Matching Model.}
$D_{kg}=\{(h_i,k_i,l_i)\}^N_{i=1}$, where $N$ means there are $N$ pieces of training data, $h_i= \{u_{i,1} ,...,u_{i,n}\}$ represents a conversation context with $u_{i,n}$ as a utterance, $k_i$ is the knowledge sampled in the ratio of positive and negative samples 1:4.
And $l_i$ indicates the corresponding label, if $k_i$ is the relevant knowledge to be replied then it is 1, otherwise, it is 0.

\noindent
\textbf{Conversation Generation Model.}
$D_{cg}=\{(h_i,k_i,r_i,p_i)\}^N_{i=1}$, where $N$ means there are $N$ pieces of training data, $h_i= \{u_{i,1} ,...,u_{i,n}\}$ represents a conversation context with $u_{i,n}$ as a utterance, $k_i$ is the knowledge.
And $r_i$ indicates the corresponding reply sampled in the ratio of positive and negative samples 1:1.
$p_i$ indicates the label of whether the $r_i$ is the positive sample, if $p_i$  is the correct reply then it is 1, otherwise, it is 0.
The positive and negative sample responses $r_i$ and $p_i$ are constructed for multi-task training, see section 3.6 for details.

\subsubsection{Inference Phase}
At this phase, we have only dialogue history and knowledge, just as we talk in reality.
Different models are required to work together to generate the corresponding response $r_i$ using datasets $D_{ip}$ and $K_{kc}$, where $D_{ip}= \{u_1 ,...,u_n\}$ represents a conversation context with $u_n$ as a utterance.
And response $r_i$ is then incorporated into dataset $D_{ip}$ for subsequent generation.

\subsection{System Architecture}

\begin{figure*}[h]
    \centering
    \centerline{
        \includegraphics[width=1\linewidth]{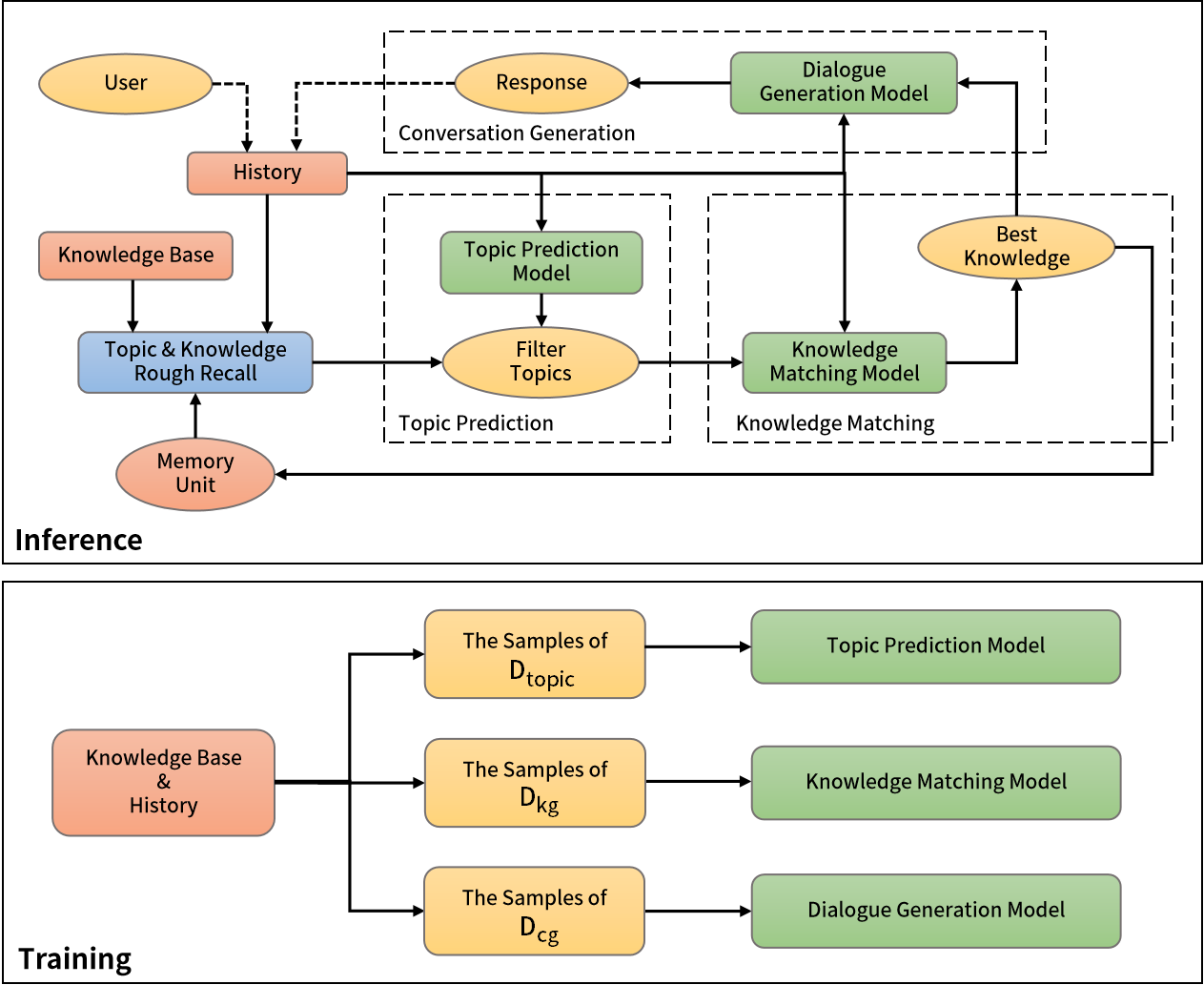}}
    \caption{System Architecture. 
In the training phase, we sample the data from KdConv and reconstruct the samples needed for training, as shown in Section 3.1.1.
The reconstructed data is then used to train the modules separately.
In the inference phase, the utterance of user and model response are packaged into a historical corpus, which is processed collaboratively by Topic and Knowledge Rough Recall algorithm, Topic Prediction Model, and Knowledge Matching Model to find the best knowledge before being sent to the Dialogue Generative Model for sentence response generation.
}
\label{fig:System Architecture}
\end{figure*}

The architecture of our system is shown in Figure \ref{fig:System Architecture}.
The training and inference processes are different: For training, we only need to use the sampled reconstructed data from KdConv for model training. Inference, on the other hand, requires multiple models to work together.

\par
In the inference phase, when the user says a sentence, the system first concatenates it behind the historical utterances.
And the system conducts a rough recall of the topics based on the historical utterances, then use these topics and memory unit to roughly recall the corresponding knowledge in the knowledge base.
The Topic Prediction Model sorts topics based on the historical utterances, and compare the results with the topics of the rough recall in the previous step, and select the most suitable topic.
Next, it input the knowledge consistent with the best topic in rough recall and the historical utterances into the Knowledge Matching Model, and rank the knowledge to get the best knowledge.
Finally, the best knowledge and historical utterances are sent to the Generative Model to generate a reply, which will be added to the end of the historical utterances.

\subsection{Rough Recall}
The conversation system needs to generate replies quickly.
In the pre-training era, if every piece of knowledge is sent to the deep learning model for processing, the response speed can not meet the conversation requirement.
So we propose to use mature and fast algorithms to roughly recall topics and knowledge.
We will compare the three algorithms of TF-IDF, LAC, and Aho-Corasick.

\noindent
\textbf{TF-IDF.}
TF-IDF is a text keyword extraction algorithm. 
Here we use the TF-IDF tool provided by jieba\footnote{\url{https://github.com/fxsjy/jieba}}.

\noindent
\textbf{LAC\footnote{\url{https://github.com/baidu/lac}}.}
The full name of LAC is Lexical Analysis of Chinese, which is a joint lexical analysis tool developed by Baidu's Natural Language Processing Department to realize Chinese word segmentation, part-of-speech tagging, proper name recognition and other functions.
In this paper, we use LAC tools for part-of-speech tagging and entity recognition.
LAC tool supports importing self-built dictionary for searching.

\noindent
\textbf{Aho-Corasick.}
The Aho-Corasick algorithm is a classic algorithm in multi-pattern matching and is currently used in many practical applications.
Here we use the ahocorasick\footnote{\url{https://github.com/WojciechMula/pyahocorasick}} tool to perform the Aho-Corasick algorithm.

We assume that $T_0$ is the result set of the topics and $K_{0}$ is the related knowledge set of $T_0$.
One of the topics as the root node is related to multiple pieces of knowledge.
As the root node, `$In\ the\ Name\ of\ the\ Father$' is connected to a lot of knowledge, as shown in Figure \ref{fig:Knowledge Graph}.
So in the next steps, we also need to recall the related topics and knowledge.

\subsection{Topic Prediction}
\begin{figure*}[h!]
\centering
\includegraphics[width=0.8\linewidth]{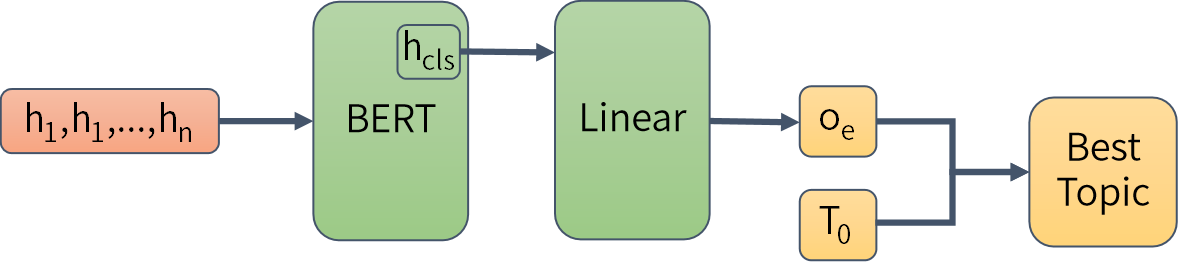}
\caption{Topic Prediction. The Topic Prediction Model is a multi-classification model that can output all topic scores $O_e$.
We sort the topics according to the value of the score and then choose the topic with the highest score in $T_0$ as the best knowledge.}
\label{fig:Topic Prediction}
\end{figure*}

The Topic Prediction Model needs to output the probability value of each topic in the knowledge base based on historical dialogue, as shown in Figure \ref{fig:Topic Prediction}.
Here we use the pre-trained model RoBERTa-wwm-ext\footnote{\url{https://github.com/ymcui/Chinese-BERT-wwm}} \cite{cui2019pre} for fine-tuning.
We put the [CLS] hidden state vector into a linear layer which outputs the final classification results $O_e$ whose dimensions are the sum of the number of topics:
$$
O_{e}=\operatorname{softmax}\left(\operatorname{linear}\left(O_{bert}\right)\right),
$$
where $O_{bert}$ are the outputs of the RoBERTa-wwm-ext.
In order to calculate faster, we can also directly use the outputs of the linear layer as the results $O_e$ without using the softmax function.
\par
Since $O_e$ is the probability value of all topics, we only need to find the topic with the largest probability value at the same time it is also in $T_0$, which is the best topic.
When we select the best topic, we need to filter out the knowledge that is not related to it in $K_{0}$, and the remaining knowledge related to the best topic is the set $K_{1}$.

\subsection{Knowledge Matching}
\textbf{Sentence-Bert} \cite{DBLP:conf/emnlp/ReimersG19}. We use Sentence-Bert as the knowledge Matching Model and use pre-trained RoBERTa-wwm-ext to initialize the Bert \cite{DBLP:conf/naacl/DevlinCLT19} module inside the model.
Because some historical data or knowledge is too long, a mount of data will be truncated and the calculation time will be too much when they are sent to RoBERTa-wwm-ext after splicing.
Sentence-Bert can greatly alleviate this situation. The twin BERT of the Sentence-Bert is used to process history and knowledge separately, it can accommodate longer data and reduce calculation time.

\begin{figure*}[h!]
\centering
\includegraphics[width=0.5\linewidth]{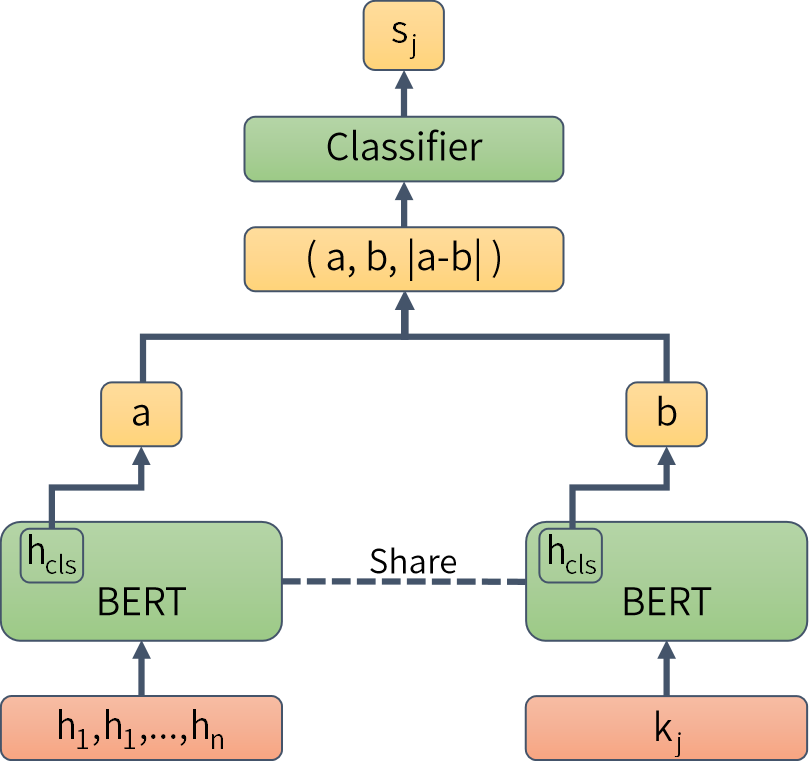}
\caption{Sentence-Bert. Knowledge Matching Model is a two-classification model that can output a knowledge score $s_j$.
This score measures how well the knowledge $k_j$ matches the histories ${h_1,..,h_n}$ of the conversation.}
\label{fig:Knowledge Matching}
\end{figure*}

As shown in Figure \ref{fig:Knowledge Matching}, the knowledge Matching Model is a two-class model.
We respectively encode the historical utterances and knowledge by the Bert and use the [CLS] hidden states vector as vector $a$ and vector $b$ respectively.
Then perform the corresponding splicing calculation operation on vector $a$ and vector $b$, and put the result vector to a linear layer which outputs the final classification results $S_j$.
We train the model for ranking knowledge by optimizing the cross-entropy loss: 
$$
\mathcal{L}_{sbert}=-\sum_{j \in J_{\mathrm{pos}}} \log \left(s_{j}\right)-\sum_{j \in J_{\mathrm{neg}}} \log \left(1-s_{j}\right),
$$
where $s_{j}$ is a score for each candidate knowledge in $K_{1}$ independently. $J_{pos}$ is the set of indexes of the appropriate candidate knowledge and $J_{neg}$ is the set of indexes of the non-appropriate candidate knowledge in $K_{1}$.
We sort all the candidate knowledge in set $K_{1}$ according to the scores $s_{j}$ and then select the first $n$ pieces of knowledge in the ranking result for dialogue generation.

\noindent
\textbf{Pairwise Model}.
We also try the pairwise ranking strategy \cite{DBLP:journals/corr/abs-1910-14424} to expect a better result.
The Pairwise Model model is the same with Topic Prediction Model, but the difference is the input data: We input up to three historical utterances together with one positive and one negative knowledge samples into the model.
We trained the model with the following loss:
$$
\begin{aligned}
\mathcal{L}_{\mathrm{ranking}}=-& \sum_{i \in J_{\mathrm{pos}}, j \in J_{\mathrm{neg}}} \log \left(s_{i, j}\right) \\
&-\sum_{i \in J_{\mathrm{neg}}, j \in J_{\mathrm{pos}}} \log \left(1-s_{i, j}\right),
\end{aligned}
$$
where $s_{i,j}$ is a score for each pair candidate knowledge in $K_{1}$. $J_{pos}$ is the set of indexes of the relevant candidate knowledge and $J_{neg}$ is the set of indexes of the non-relevant candidate knowledge in $K_{1}$.

\subsection{Conversation Generation}
At the core of our approach is language modeling \cite{DBLP:journals/jmlr/BengioDVJ03}.
We first concatenate the utterances in a multi-turn dialogue session into $N-1$ samples ($N$ is the number of utterances).
Each sample is concatenated into a long text $S= k_{n}, s_1, ... , s_{n-1}$ ($s_i$ is the historical utterances of the current reply where $i \in [1, n-1]$.
$k_{n}$ is the response-related knowledge.) by the dialog history and response-related knowledge.
And we denote the target sentence (ground truth response) as $s_n$, the conditional probability of $P(s_{n}|S)$ can be written as the product of a series of conditional probabilities:
$$
p(s_n \mid S)= p\left(s_{n} \mid k_n, s_{1}, \cdots, s_{n-1}\right),
$$
where, $n\in [2, N]$. The generation condition probability of each token in $s_n$ is $P(w_{i}|S, w_{1},...w_{i-1})$, and the generation condition probability of $P(s_{n})$ is as follows :
$$
p(s_n)=\prod_{i=1}^{m} p\left(w_{i} \mid S, w_{1}, \cdots, w_{i-1}\right),
$$
where, $m$ is the number of tokens in $s_n$.

\begin{figure*}[h!]
\centering
\centerline{
\includegraphics[width=0.8\linewidth]{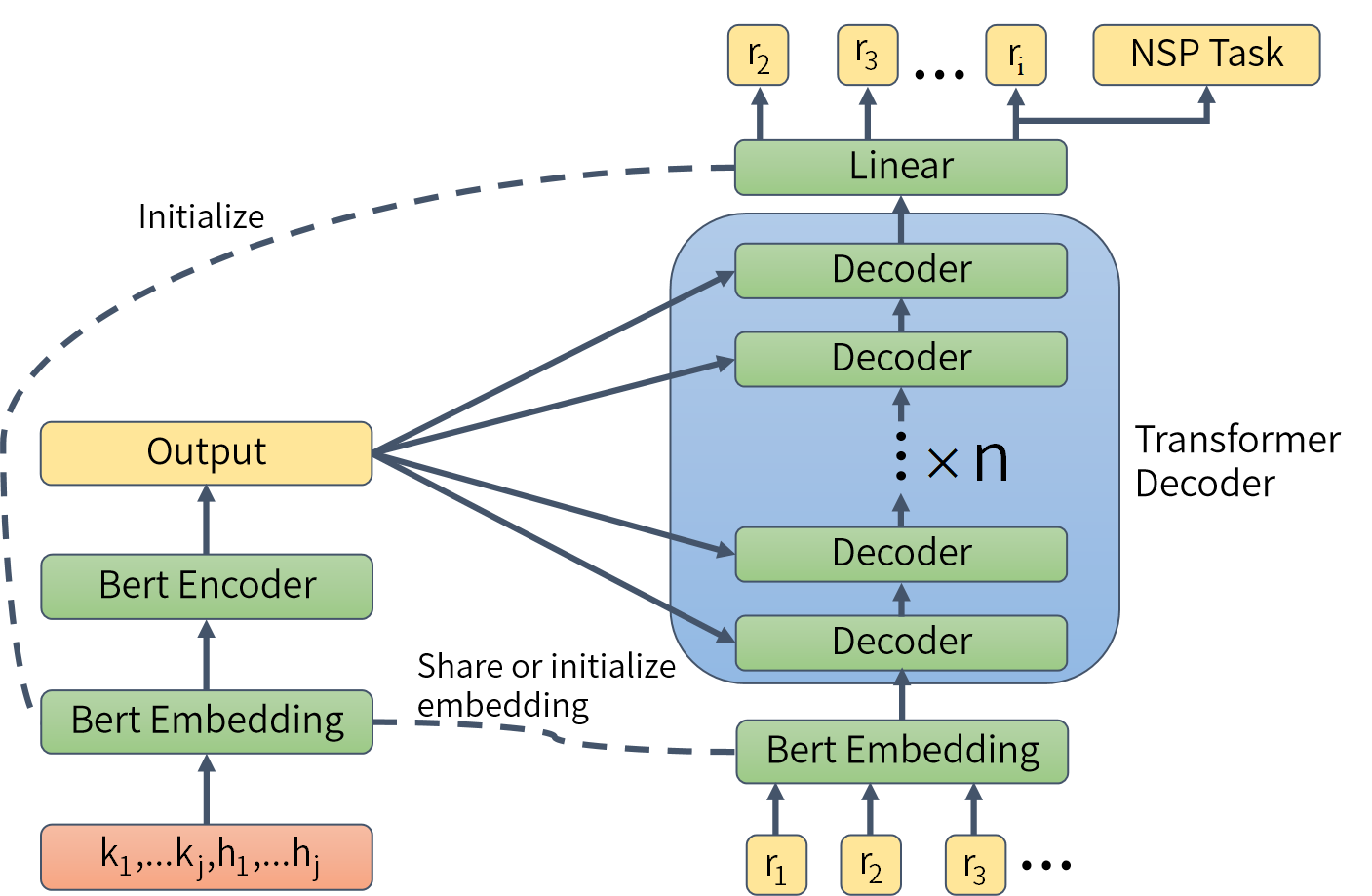}}
\caption{Bert2Transformer. The encoder is Bert base and the decoder is 12-layers Transformer's decoder. And using Bert's embedding to initialize the embedding and output linear layer of the decoder.
Training this model through multi\_task: LM task and take the last token as a binary classification task(NSP task).
}
\label{fig:Bert2Transformer}
\end{figure*}

\begin{figure*}[h!]
\centering
\includegraphics[width=0.7\linewidth]{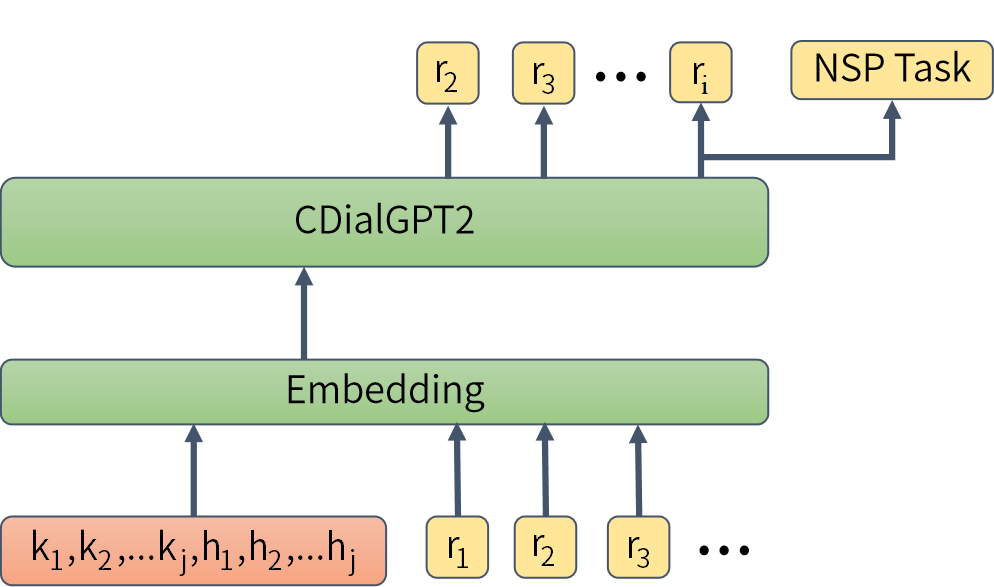}
\caption{CDialGPT2. CDialGPT2 is a 12-layer GPT2, we fine-tuning this model by multi\_task: LM task and take the last token as a binary classification task (NSP task)}
\label{fig:CDialGPT2}
\end{figure*}

\par
In order to explore the effect of different models using knowledge generation, we use Bert2Transformer and CDial-GPT \cite{DBLP:conf/acl/ZhangSGCBGGLD20} as the generation models.
At the same time, in the training phase, we designed a multi-task what is similar to the NSP in Bert \cite{DBLP:conf/naacl/DevlinCLT19}, hoping to allow the generation model to learn to distinguish sentences that are close in literal distance but farther away in the semantic distance. eg:
Suppose our context is talking about dogs, the model may generate one of the following two sentences: 
[`Hello, my dog is cute.', `Hello, my cat is cute.'].
For the model, the generation probabilities of these two sentences are too close and maybe generate incorrectly.
So we construct a multi-task: a two-classification task, let the model judge whether the current reply is a suitable reply.
As shown in Figure \ref{fig:Bert2Transformer} and Figure \ref{fig:CDialGPT2}, because the last token of the generation task have global information, we take the hidden states vector of the last token from the reply into the linear layer and activation function and then output the logits as the binary-classification result.
\par
The $\mathcal{L}_{total}$ is calculated as follows:
$$
\mathcal{L}_{total} = \alpha\mathcal{L}_{\textrm{LM}} + (1-\alpha)\mathcal{L}_{\textrm{NSP}},
$$
where $\mathcal{L}_{total}$ is the total loss of training, $\mathcal{L}_{\textrm{LM}}$ is the language model's loss and the $\mathcal{L}_{\textrm{NSP}}$ is the MultitaskNSP's loss.
In this experiment, if the reply in the current training sample is suitable, $\alpha$ is set as 0.5.
Otherwise, $\alpha$ is set as 0.

\noindent
\textbf{Bert2Transformer.}
As shown in Figure \ref{fig:Bert2Transformer}, the Bert2Transformer model is based on the Transformer \cite{DBLP:conf/nips/VaswaniSPUJGKP17} framework, but the encoder is Bert base and the decoder is a 12-layers Transformer's decoder.
In order for the decoder to obtain some prior information, we use Bert's embedding to initialize both the embedding and output linear layer of the decoder.
At the same time, we compare whether the embedding of the encoder and the decoder share parameters so as to affect the generation effect.
At the same time, compare the different performance of whether the encoder embedding layer and the decode embedding layer share parameters.

\noindent
\textbf{CDial-GPT2.}
As shown in Figure \ref{fig:CDialGPT2}, CDialGPT2 is a 12-layer GPT2 which is pre-trained for 70 epochs on the Chinese novel dataset \cite{DBLP:conf/nlpcc/WangKZHJZH20} and post-trained for 30 epochs on LCCC-base \cite{DBLP:conf/nlpcc/WangKZHJZH20}.

\section{Experiments}
All models are based on the transformers\footnote{\url{https://github.com/huggingface/transformers}} framework and use pre-trained models RoBERTa-wwm-ext, RoBERTa-wwm-ext-large\footnote{\url{https://github.com/ymcui/Chinese-BERT-wwm}} and CDial-GPT2\_LCCC-base\footnote{\url{https://github.com/thu-coai/CDial-GPT}} to fine-tuning respectively.
And using CrossEntropyLoss as the loss function for training.
At the same time, all deep learning models are optimized by the AdamW \cite{DBLP:journals/corr/abs-1711-05101} optimizer and the learning rate decay method.
We create a schedule with a learning rate that decreases linearly from the initial learning rate set in the optimizer to 0.

\subsection{Data Process}
\begin{figure*}[h!]
\centering
\centerline{
\includegraphics[width=1\linewidth]{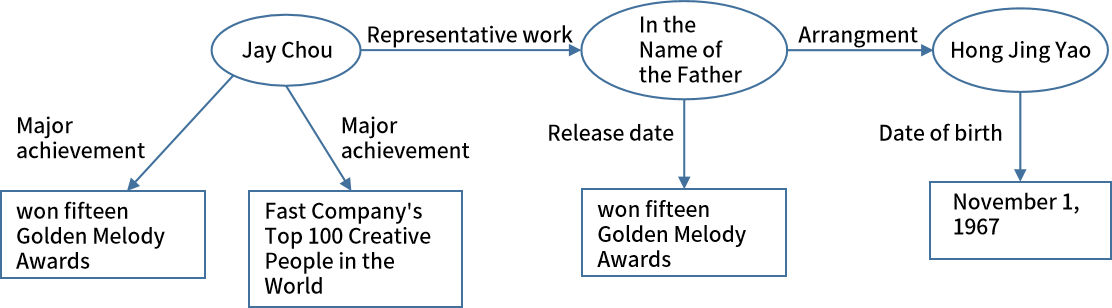}}
\caption{Knowledge Graph.}
\label{fig:Knowledge Graph}
\end{figure*}

We use the KdConv\footnote{\url{https://github.com/thu-coai/KdConv}} \cite{DBLP:conf/acl/ZhouZHHZ20} for experiments. KdConv is a Chinese multi-domain knowledge-driven conversation dataset, which grounds the topics in multi-turn conversations to knowledge graphs. 
The corpus contains 4.5K conversations from three domains (film, music, and travel), and 86K utterances with an average turn number of 19.0.
In order to explore the factors that may affect the generation of the reply, we need to process and structure the dataset into the format required by chapter 3.1. 
As for the knowledge graph, as shown in Figure \ref{fig:Knowledge Graph}, we just need to read in all of it and search when needed.

For all training models, the input data starts with [CLS] token and ends with [SEP] token.

\subsection{Evaluation Metrics}

On all the dataset we use average BLEU \cite{DBLP:conf/acl/PapineniRWZ02} score for 1, 2, 3, 4-gram and Distinct-2 \cite{DBLP:conf/naacl/LiGBGD16} to measure the final generation quality. 
We use BLEU to measure the similarity between the generated sentences and labels.
Distinct measures the degree of diversity by calculating the number of distinct unigrams and bigrams in generated responses.

\subsection{Rough Recall Implementation Details and Results}
In the Rough Recall stage, we compare TF-IDF, LAC and Aho-Corasick algorithms respectively.
Because they are all ready-made tools, we directly use all the training data as the test samples for testing.
For each sample, we can take up to the last 10 sentences of dialogue history as the historical utterances.

\noindent
\textbf{TF-IDF.}
We take all the knowledge root nodes in the knowledge graph as topics and add them to the TF-IDF dictionary so that the jieba tool can correctly cut the knowledge nodes during word segmentation.
After cutting the historical dialogue utterances with jieba and removing the stop words, they are sent to the TF-IDF algorithm, and the first $n$ results of the outputs are taken as the result set.

\noindent
\textbf{LAC.}
We take all the knowledge root nodes in the knowledge graph as topics and add them to the LAC dictionary so that the LAC algorithm can accurately identify the topics.
Because the results of the LAC algorithm are generated in the order of the input utterances, we take the last $n$ results generated from LAC as the result set.

\noindent
\textbf{Aho-Corasick.}
We take the Automaton class of ahocorasick algorithm as a trie and add the root node of knowledge considered as a topic as well as its associated values to this trie.
In the test, we input the historical utterances, take all the output results, sort them by string length from largest to smallest, and take the top $n$ results as the result set.

\par
We use accuracy as a measure to compare the accuracy of the first $n\in[1, 50]$ outputs of these three algorithms that contain the correct topics, in other words, each algorithm tests the accuracy of 50 different values.

Assuming there are $T$ evaluation samples, in this experiment $T=62938$, the specific formula is as follows:
$$
\begin{array}{l}
y_{it}=\left\{\begin{array}{l}
1, \quad\text { if }\quad k_{t} \in O_{it}, 
\\
0, \quad\text { if }\quad k_{t} \notin O_{it},
\end{array}\right.
\\
\\
acc_{i}={\sum_{t=1}^{T} y_{it}}/{T} \times 100 \%,
\end{array}
$$
where $i\in(0,1,2)$ respectively indicates which algorithms of TF-IDF, LAC and Aho-Corasick is used, $k_{t}$ represents the true topic of the $t$-th sample, $O_{it}$ represents the result set predicted by the $i$-th algorithm for the $t$-th sample, $y_{it}$ is the score calculated on the $t$-th sample using the $i$-th algorithm, and $acc_{i}$ is the final accuracy rate of the $i$-th algorithm.
\par
\begin{figure*}[h!]
\centering
\includegraphics[width=1\linewidth]{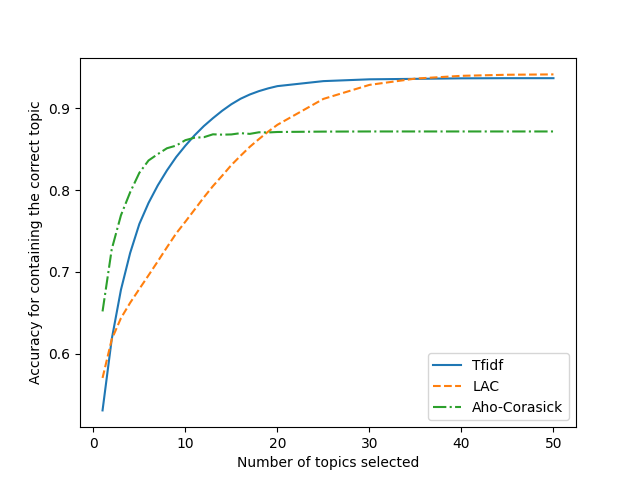}
\caption{Rough Recall Results.
The abscissa represents the number of topics recalled by different algorithms, and the ordinate represents the accuracy of different algorithms in recalling that number of topics. 
When selecting the first 50 topics, the accuracy rate of LAC can reach 94\%.}
\label{fig:theme_acc}
\end{figure*}

The results of different coarse recall algorithms are as follows Figure \ref{fig:theme_acc}.
In the case of selecting different numbers of topics, different algorithms perform differently.
The LAC algorithm performs best when the number of recalled topics is high, and its accuracy can reach 94\% when selecting the first 50 topics.
In the process of matching the topics of coarse recall with the results predicted by the Topic Prediction Model, we hope that the accuracy the higher, the better. 
So in the next experimental process, we choose LAC as our coarse recall algorithm.

\subsection{Topic Prediction Model Training Details and Results}
Since there are 12,149 topics, the topic prediction model is a multi-classification model with a total of 12,149 classification results.
We use historical utterances as input data and the node element of the triad in the knowledge graph is the classification label, as shown in Figure \ref{fig:Knowledge Graph}.
Among them, the longest input data length is truncated to 400, and the last 10 sentences of the historical corpus are used at most.
At the same time, using [SEP] token to separate different dialogue utterances.
Then we experiment with the base model and the large model separately.

\noindent
\textbf{Base Model.}
Fine-tuning through RoBERTa-wwm-ext.
The learning rate is initialized to 2e-5.
The batch size is set to 14.

\noindent
\textbf{Large Model.}
Fine-tuning through RoBERTa-wwm-ext-large.
The learning rate is initialized to 1e-5.
The batch size is set to 2.
\par
The results are shown in Table \ref{table:Topic Prediction Model evaluation}.
It can be seen that the large model is slightly better than the base model in performance.

\begin{table}[htb]      
\centering
\caption{Topic Prediction Model evaluation.
Accuracy here refers to the accuracy with which the model correctly predicts the topic, and it measures the performance of the model in recalling historical conversation topics.}
\label{tab1}
    \begin{tabular}{ccc}    
        \toprule      
        \textbf{Model} & \textbf{Valid Accuracy} & \textbf{Test Accuracy} \\        
        \midrule            
        {RoBERTa-wwm-ext} & {90.38\%} & {85.23\%} \\        
        {RoBERTa-wwm-ext-large} & {91.19\%} & {86.06\%} \\
        \bottomrule         
    \end{tabular}
    \label{table:Topic Prediction Model evaluation}
\end{table}

\subsection{Knowledge Matching Model Training Details and Results}
The knowledge Matching Model is a two-classification model, which inputs knowledge and dialogue historical utterances, and outputs a logits value or probability value.
The longest historical utterances data length and knowledge length are respectively controlled within 400, and at most the last 10 sentences of the historical corpus are used as historical utterances data.
We use [SEP] token to separate different dialogue utterances.
In the experiment, we compare Sentence-Bert model and pairwise model separately.
Since Sentence-Bert needs to encode historical statements and knowledge data separately, we also test the effect after encoding the data using the twin Bert and two different Bert.

\noindent
\textbf{Sentence-Bert Base.}
Fine-tuning through RoBERTa-wwm-ext.
The learning rate is initialized to 3e-5.
The batch size was set to 14.

\noindent
\textbf{Sentence-Bert Large.}
Fine-tuning through RoBERTa-wwm-ext-large.
The learning rate is initialized to 1e-5.
The batch size was set to 2.

\noindent
\textbf{Pairwise Model.}
We fine-tuning this model through RoBERTa-wwm-ext.
The maximum input data length is controlled within 500.
And the learning rate is initialized to 2e-5.
The batch size was set to 14.

The results are shown in Table \ref{table:knowledge Matching Model evaluation}.
We compare the accuracy of different models on the validation set and test set.
The accuracy here is only the accuracy of the model's binary classification of data, which is different from the accuracy of the final knowledge selection.

From the results, the performance of using twin Bert is much better than using two different Bert.
This is maybe because two Bert are used to separately encode the knowledge and historical utterances, so they only learn part of the data, but not the other side's data, and the data lacks sufficient interaction.

\begin{table}[htb]      
\centering
\caption{knowledge Matching Model evaluation.
The accuracy is only the accuracy of the model to correctly classify the data, not the knowledge selection accuracy."-diff" means using two different Bert to encoding knowledge and historical utterances.}
\label{tab1}
    \begin{tabular}{ccc}    
        \toprule      
        \textbf{Model} & \textbf{Valid Accuracy} & \textbf{Test Accuracy} \\        
        \midrule            
        {Sentence-Bert+base} & {96.09\%} & {95.54\%} \\ 
        {Sentence-Bert+base+diff} & {94.75\%} & {73.49\%} \\ 
        {Sentence-Bert+large} & {96.14\%} & {95.38\%} \\
        {Pairwise Model} & {90.12\%} & {89.64\%} \\
        \bottomrule         
        \end{tabular}
        \label{table:knowledge Matching Model evaluation}
\end{table}

The accuracy of the final knowledge selection is shown in Table \ref{table:Knowledge Selection Accuracy}.
This shows the accuracy from rough recall to final knowledge selection.
The rough recall stage we use the LAC algorithm, input up to the last 10 historical utterances, and output 50 topics for subsequent model processing.
We compare the accuracy of the first knowledge being appropriate knowledge and the accuracy of the first three knowledge and the first five knowledge containing appropriate knowledge separately after sorting the knowledge.
\par
From the results, the Pairwise Model is not suitable for knowledge screening.
Our intention is to allow the model to pay more attention to the interaction between positive sample knowledge and negative sample knowledge, so as to determine more appropriate knowledge.
But judging from the results, the added knowledge seems to become noise, which interferes with model judgment.
For Sentence-bert, the model is more concerned with an alignment between knowledge and dialogue history.
The better the aligned knowledge, the more appropriate the model thought it is.

\begin{table}[htb]
\centering
\caption{Knowledge Selection Accuracy.
This shows the accuracy from rough recall to final knowledge selection.
`No.' indicates the number of selected knowledge. 
And we respectively show the accuracy of correct knowledge contained in the first 1, 3, and 5 pieces of knowledge after sorting.
`\Checkmark' indicates the fine-tuned pre-training model used in the current experiment.
`$\varnothing$' means not to use or not test.
}
\resizebox{\textwidth}{17mm}{
\begin{tabular}{cc|cc|c|ccc}
\toprule
\multicolumn{2}{c|}{\textbf{Topic Prediction Model}} & \multicolumn{2}{c|}{\textbf{Sentence-Bert}} & \multirow{2}{*}{\begin{tabular}[c]{@{}c@{}}\textbf{Pairwise Model} \\ \textbf{Model}\end{tabular}} & \multicolumn{3}{c}{\textbf{No.}} \\ \cline{1-4} \cline{6-8} 
\textbf{base} & \textbf{large} & \textbf{base} & \textbf{large} & & \textbf{1} & \textbf{3} & \textbf{5} \\ 
\hline
\Checkmark & $\varnothing$ 	& \Checkmark & $\varnothing$    & $\varnothing$  	& 28.66\% & 49.12\% & 60.25\% \\
\Checkmark & $\varnothing$ 	& $\varnothing$    & \Checkmark & $\varnothing$  	& 21.86\% & 47.03\% & 59.23\% \\
$\varnothing$    & \Checkmark & \Checkmark & $\varnothing$    & $\varnothing$  	& \underline{28.94\%} & \underline{49.80\%} & \underline{61.05\%} \\
$\varnothing$    & \Checkmark & $\varnothing$    & \Checkmark & $\varnothing$  	& 22.10\% & 47.59\% & 59.83\%	\\
\Checkmark & $\varnothing$    & $\varnothing$    & $\varnothing$    & \Checkmark 	& 9.92\%  & $\varnothing$ & $\varnothing$\\
$\varnothing$    & \Checkmark & $\varnothing$    & $\varnothing$    & \Checkmark  & 10.06\% & $\varnothing$ & $\varnothing$ \\
\bottomrule
\end{tabular}}
\label{table:Knowledge Selection Accuracy}
\end{table}

\subsection{Conversation Generation Model Training Details and Results}

\begin{table}[htp]
\small
\centering
\caption{Generation Results.
`Seq2Seq' and `HRED' come from the baseline of the corresponding Generative Model in the dataset KdConv.
`+ NSP' means the model trained by multitasking, refer to Chapter 4.6 `MultitaskNSP'.
`+ 1kb' indicates the model contains one of knowledge per training sample during training;
`+ 3kb' indicates the model contains three of knowledge per training sample during training;
`+ share' indicates encoder share embedding with decoder;
`AVG.B' denotes average BLEU score for 1, 2, 3, 4-gram.
`Dis-2' denotes for Distinct-2.
}
\begin{tabular}{ccccc}	

\toprule
\multicolumn{1}{c|}{\multirow{2}{*}{\textbf{Model}}} & \multicolumn{2}{c|}{\textbf{Valid}} & \multicolumn{2}{c}{\textbf{Test}} \\ 
\cline{2-5} 
\multicolumn{1}{c|}{} & \textbf{AVG.B} & \multicolumn{1}{c|}{\textbf{Dis-2}} & \textbf{AVG.B} & \textbf{Dis-2} \\ 

\midrule
\multicolumn{5}{c}{\textbf{Use true knowledge to generate}} \\ 
\midrule
\multicolumn{1}{c|}{\textbf{Seq2Seq + know}}        	   	   & $\varnothing$ & $\varnothing$ & 18.95 & 11.32 \\
\multicolumn{1}{c|}{\textbf{HRED + know}}        	   		   & $\varnothing$ & $\varnothing$ & 18.87 & 11.03 \\
\multicolumn{1}{c|}{\textbf{CDial-GPT2 + 1kb}}        	   & 28.55 & 13.22 & 27.51 & 12.22 \\
\multicolumn{1}{c|}{\textbf{CDial-GPT2 + 3kb}}        	   & 29.79 & 14.07 & 28.94 & 12.83 \\
\multicolumn{1}{c|}{\textbf{CDial-GPT2 + 1kb + NSP}}   	   & 26.76 & 12.35 & 25.35 & 11.26 \\
\multicolumn{1}{c|}{\textbf{CDial-GPT2 + 3kb + NSP}}   	   & 24.61 & 13.97 & 23.35 & 13.23 \\ 
\multicolumn{1}{c|}{\textbf{Bert2Transformer + 1kb}}   	   & 35.07 & 18.21 & 35.16 & 16.40 \\ 
\multicolumn{1}{c|}{\textbf{Bert2Transformer + 3kb}}   	   & \underline{35.74} & 18.10 & \underline{35.92} & 16.32 \\ 
\multicolumn{1}{c|}{\textbf{Bert2Transformer + 1kb + share}} & 35.36 & \underline{18.56} & 35.10 & \underline{16.61} \\ 
\multicolumn{1}{c|}{\textbf{Bert2Transformer + 3kb + share}} & 34.92 & 17.74 & 35.10 & \underline{16.61} \\ 

\midrule
\multicolumn{5}{c}{\textbf{Use the knowledge of a recall to generate}} \\ 
\midrule
\multicolumn{1}{c|}{\textbf{CDial-GPT2 + 1kb}}        	   & 15.86 & 12.10 & 15.34 & 11 17 \\
\multicolumn{1}{c|}{\textbf{CDial-GPT2 + 3kb}}        	   & 15.52 & 12.29 & 15.09 & 11.49 \\
\multicolumn{1}{c|}{\textbf{CDial-GPT2 + 1kb + NSP}}   	   & 14.72 & 10.83 & 13.93 & 10.43 \\
\multicolumn{1}{c|}{\textbf{CDial-GPT2 + 3kb + NSP}}   	   & 14.94 & 12.67 & 14.00 & 12.19 \\ 
\multicolumn{1}{c|}{\textbf{Bert2Transformer + 1kb}}   	   & 20.00 & 16.52 & 19.59 & 15.52 \\ 
\multicolumn{1}{c|}{\textbf{Bert2Transformer + 3kb}}   	   & 19.57 & 16.96 & 18.88 & 15.82 \\ 
\multicolumn{1}{c|}{\textbf{Bert2Transformer + share + 1kb}} & 20.06 & 16.56 & 19.65 & 15.75 \\ 
\multicolumn{1}{c|}{\textbf{Bert2Transformer + share + 3kb}} & 19.88 & 16.53 & 19.22 & 15 21 \\ 

\midrule
\multicolumn{5}{c}{\textbf{Use the knowledge of three recalls to generate}} \\ 
\midrule
\multicolumn{1}{c|}{\textbf{CDial-GPT2 + 3kb}}        	   & 14.79 & 13.67 & 13.87 & 13.03 \\
\multicolumn{1}{c|}{\textbf{CDial-GPT2 + 3kb + NSP}}   	   & 14.81 & 13.78 & 13.28 & 13.40 \\ 
\multicolumn{1}{c|}{\textbf{Bert2Transformer + 3kb}}   	   & \underline{23.71} & \underline{17.17} & \underline{23.24} & \underline{15.93} \\ 
\multicolumn{1}{c|}{\textbf{Bert2Transformer + share + 3kb}} & 23.52 & 16.74 & 22.75 & 15.74 \\ 

\bottomrule
\end{tabular}
\label{table:Generation Results}
\end{table}

We use CDial-GPT2 and Bert2transformer as generation models for our experiments.
Using the last 10 historical utterances and $m$ pieces of knowledge as input to generate a reply where $m=1,3$, to explore the influence of using different knowledge numbers on generation.
The longest data length of the sum of historical utterances data length and knowledge length is controlled within 400.
Using [speaker1] [speaker2] tokens to separate different historical utterances and [SEP] token to separate different knowledge.

\noindent
\textbf{CDial-GPT2.}
Fine-tuning through CDial-GPT2\_LCCC-base.
The learning rate is initialized to 3e-5.
The batch size is set to 6.

\noindent
\textbf{Bert2transformer.} Fine-tuning through RoBERTa-wwm-ext-base.
We use the Seq2Seq architecture, where RoBERTa-wwm-ext-base is used as the encoder to encode the input data, and the transformer as the decoder to generate the response.
The learning rate is initialized to 1e-5.
The batch size is set to 2.

\noindent
\textbf{MultitaskNSP.}
If CDial-GPT2 or Bert2transformer add MultitaskNSP task, set the batch size to 2 respectively, and other parameters are the same as the above settings.

In the response generation stage, the Topic Prediction Model and the Knowledge Matching Model use the large model and the base model respectively. 
We experiment with two types of models based on CDial-GPT2 and Bert2transformer, and the results are shown in the following Table \ref{table:Generation Results}.
Here we show the results generated using real knowledge, recall of one knowledge, and recall of three knowledge respectively.
Where for the knowledge recall stage we use the method with the best results in Table \ref{table:Knowledge Selection Accuracy}.

\par
From the overall results, the Bert2transformer model has the best performance.
Bert2Transformer can achieve SOTA when training and generating with three pieces of knowledge.
And multi-task training does not improve model performance but decrease.
\par
From the results generated using the true knowledge, the model that selects three pieces of knowledge for training has greater potential than the model that selects one piece of knowledge for training.
We speculate that this is because using multiple knowledge during training is equivalent to adding noise input to the model, enhancing the robustness of the model.
\par
From the results of using recall knowledge to generate, when using multiple knowledge to train and generate, the performance of CDial-GPT2 is reduced, but the performance of Bert2transformer is improved.
Extra knowledge turn into noise, which affect the generation effect of CDial-GPT2.
But Bert2transformer can select the most appropriate knowledge among multiple knowledge for generating.
It can be seen that CDial-GPT2 is not as good as the Bert2transformer for extracting key information.

\section{Conclusion}
Knowledge-driven Conversation is very important because it makes the generated text more like humans.
In this paper, we propose a knowledge-driven conversation system and a dialogue generation model Bert2transformer.
The knowledge-driven conversation system includes three modules such as topic prediction, knowledge matching, and dialogue generation.
Based on this system and the KdConv dataset, we explore the key factors that influence the task of generating knowledge-driven conversation:
For coarse recall, using the LAC algorithm to recall more topics can improve the accuracy of the system's topic prediction, in this paper we reach 94\%.
For the topic prediction phase, our model achieves 91.19\% and 86.06\% accuracy in predicting topics on the validation and test sets, respectively.
In the knowledge matching phase, our model achieves 96.14\% and 95.38\% accuracy on the validation and test sets, respectively.
When ranking the knowledge, our model achieves 61.05\% accuracy.
Finally, in the conversation generation phase, our Bert2transformer model outperforms both baseline and CDial-GPT2 in terms of BLEU and Distinct results, reaching 23.24 and 15.93, respectively.

\begin{acknowledgements}
This work is supported by the Key Program of National Science Foundation of China (Grant No. 61836006) and partially supported by National Natural Science Fund for Distinguished Young Scholar (Grant No. 61625204). 
\end{acknowledgements}

\section*{Conflict of interest}
\textbf{Conflict of interest} The authors declare that they have no conflict of interest.

\bibliographystyle{spbasic}      

\begin{thebibliography}{33}
\providecommand{\natexlab}[1]{#1}
\providecommand{\url}[1]{{#1}}
\providecommand{\urlprefix}{URL }
\expandafter\ifx\csname urlstyle\endcsname\relax
  \providecommand{\doi}[1]{DOI~\discretionary{}{}{}#1}\else
  \providecommand{\doi}{DOI~\discretionary{}{}{}\begingroup
  \urlstyle{rm}\Url}\fi
\providecommand{\eprint}[2][]{\url{#2}}

\bibitem[{Bengio et~al.(2003)Bengio, Ducharme, Vincent, and
  Janvin}]{DBLP:journals/jmlr/BengioDVJ03}
Bengio Y, Ducharme R, Vincent P, Janvin C (2003) A neural probabilistic
  language model. J Mach Learn Res 3:1137--1155,
  \urlprefix\url{http://jmlr.org/papers/v3/bengio03a.html}

\bibitem[{Chen et~al.(2019)Chen, Peng, Wang, Xu, and
  Wu}]{DBLP:conf/ijcai/ChenPWXW19}
Chen C, Peng J, Wang F, Xu J, Wu H (2019) Generating multiple diverse responses
  with multi-mapping and posterior mapping selection. In: Kraus S (ed)
  Proceedings of the Twenty-Eighth International Joint Conference on Artificial
  Intelligence, {IJCAI} 2019, Macao, China, August 10-16, 2019, ijcai.org, pp
  4918--4924, \doi{10.24963/ijcai.2019/683},
  \urlprefix\url{https://doi.org/10.24963/ijcai.2019/683}

\bibitem[{Chen et~al.(2018)Chen, Zhu, Ling, Inkpen, and
  Wei}]{DBLP:conf/acl/InkpenZLCW18}
Chen Q, Zhu X, Ling Z, Inkpen D, Wei S (2018) Neural natural language inference
  models enhanced with external knowledge. In: Gurevych I, Miyao Y (eds)
  Proceedings of the 56th Annual Meeting of the Association for Computational
  Linguistics, {ACL} 2018, Melbourne, Australia, July 15-20, 2018, Volume 1:
  Long Papers, Association for Computational Linguistics, pp 2406--2417,
  \doi{10.18653/v1/P18-1224},
  \urlprefix\url{https://www.aclweb.org/anthology/P18-1224/}

\bibitem[{Cui et~al.(2019)Cui, Che, Liu, Qin, Yang, Wang, and Hu}]{cui2019pre}
Cui Y, Che W, Liu T, Qin B, Yang Z, Wang S, Hu G (2019) Pre-training with whole
  word masking for chinese bert. arXiv preprint arXiv:190608101

\bibitem[{Devlin et~al.(2019)Devlin, Chang, Lee, and
  Toutanova}]{DBLP:conf/naacl/DevlinCLT19}
Devlin J, Chang M, Lee K, Toutanova K (2019) {BERT:} pre-training of deep
  bidirectional transformers for language understanding. In: Burstein J, Doran
  C, Solorio T (eds) Proceedings of the 2019 Conference of the North American
  Chapter of the Association for Computational Linguistics: Human Language
  Technologies, {NAACL-HLT} 2019, Minneapolis, MN, USA, June 2-7, 2019, Volume
  1 (Long and Short Papers), Association for Computational Linguistics, pp
  4171--4186, \doi{10.18653/v1/n19-1423},
  \urlprefix\url{https://doi.org/10.18653/v1/n19-1423}

\bibitem[{Ghazvininejad et~al.(2018)Ghazvininejad, Brockett, Chang, Dolan, Gao,
  Yih, and Galley}]{DBLP:conf/aaai/GhazvininejadBC18}
Ghazvininejad M, Brockett C, Chang M, Dolan B, Gao J, Yih W, Galley M (2018) A
  knowledge-grounded neural conversation model. In: McIlraith SA, Weinberger KQ
  (eds) Proceedings of the Thirty-Second {AAAI} Conference on Artificial
  Intelligence, (AAAI-18), the 30th innovative Applications of Artificial
  Intelligence (IAAI-18), and the 8th {AAAI} Symposium on Educational Advances
  in Artificial Intelligence (EAAI-18), New Orleans, Louisiana, USA, February
  2-7, 2018, {AAAI} Press, pp 5110--5117,
  \urlprefix\url{https://www.aaai.org/ocs/index.php/AAAI/AAAI18/paper/view/16710}

\bibitem[{Jin et~al.(2018)Jin, Lei, Ren, Chen, Liang, Zhao, and
  Yin}]{jin2018explicit}
Jin X, Lei W, Ren Z, Chen H, Liang S, Zhao Y, Yin D (2018) Explicit state
  tracking with semi-supervisionfor neural dialogue generation. In: Proceedings
  of the 27th ACM International Conference on Information and Knowledge
  Management, pp 1403--1412

\bibitem[{Lei et~al.(2018)Lei, Jin, Kan, Ren, He, and Yin}]{lei2018sequicity}
Lei W, Jin X, Kan MY, Ren Z, He X, Yin D (2018) Sequicity: Simplifying
  task-oriented dialogue systems with single sequence-to-sequence
  architectures. In: Proceedings of the 56th Annual Meeting of the Association
  for Computational Linguistics (Volume 1: Long Papers), pp 1437--1447

\bibitem[{Lei et~al.(2020{\natexlab{a}})Lei, He, de~Rijke, and
  Chua}]{10.1145/3397271.3401419}
Lei W, He X, de~Rijke M, Chua TS (2020{\natexlab{a}}) Conversational
  recommendation: Formulation, methods, and evaluation. In: Proceedings of the
  43rd International ACM SIGIR Conference on Research and Development in
  Information Retrieval, Association for Computing Machinery, New York, NY,
  USA, SIGIR '20, p 2425–2428, \doi{10.1145/3397271.3401419},
  \urlprefix\url{https://doi.org/10.1145/3397271.3401419}

\bibitem[{Lei et~al.(2020{\natexlab{b}})Lei, Zhang, He, Miao, Wang, Chen, and
  Chua}]{10.1145/3394486.3403258}
Lei W, Zhang G, He X, Miao Y, Wang X, Chen L, Chua TS (2020{\natexlab{b}})
  Interactive path reasoning on graph for conversational recommendation. In:
  Proceedings of the 26th ACM SIGKDD International Conference on Knowledge
  Discovery \&amp; Data Mining, Association for Computing Machinery, New York,
  NY, USA, KDD '20, p 2073–2083, \doi{10.1145/3394486.3403258},
  \urlprefix\url{https://doi.org/10.1145/3394486.3403258}

\bibitem[{Li et~al.(2016{\natexlab{a}})Li, Galley, Brockett, Gao, and
  Dolan}]{DBLP:conf/naacl/LiGBGD16}
Li J, Galley M, Brockett C, Gao J, Dolan B (2016{\natexlab{a}}) A
  diversity-promoting objective function for neural conversation models. In:
  Knight K, Nenkova A, Rambow O (eds) {NAACL} {HLT} 2016, The 2016 Conference
  of the North American Chapter of the Association for Computational
  Linguistics: Human Language Technologies, San Diego California, USA, June
  12-17, 2016, The Association for Computational Linguistics, pp 110--119,
  \doi{10.18653/v1/n16-1014},
  \urlprefix\url{https://doi.org/10.18653/v1/n16-1014}

\bibitem[{Li et~al.(2016{\natexlab{b}})Li, Monroe, and Jurafsky}]{li2016simple}
Li J, Monroe W, Jurafsky D (2016{\natexlab{b}}) A simple, fast diverse decoding
  algorithm for neural generation. arXiv preprint arXiv:161108562

\bibitem[{Liang et~al.(2020)Liang, Lei, Chan, Yang, Sun, and
  Chua}]{liang2020pirhdy}
Liang H, Lei W, Chan PY, Yang Z, Sun M, Chua TS (2020) Pirhdy: Learning pitch-,
  rhythm-, and dynamics-aware embeddings for symbolic music. In: Proceedings of
  the 28th ACM International Conference on Multimedia, pp 574--582

\bibitem[{Liu et~al.(2020{\natexlab{a}})Liu, Gong, Yan, Fu, Shao, Jiang, Lv,
  and Duan}]{DBLP:conf/emnlp/LiuGYFSJLD20}
Liu D, Gong Y, Yan Y, Fu J, Shao B, Jiang D, Lv J, Duan N (2020{\natexlab{a}})
  Diverse, controllable, and keyphrase-aware: {A} corpus and method for news
  multi-headline generation. In: Webber B, Cohn T, He Y, Liu Y (eds)
  Proceedings of the 2020 Conference on Empirical Methods in Natural Language
  Processing, {EMNLP} 2020, Online, November 16-20, 2020, Association for
  Computational Linguistics, pp 6241--6250,
  \doi{10.18653/v1/2020.emnlp-main.505},
  \urlprefix\url{https://doi.org/10.18653/v1/2020.emnlp-main.505}

\bibitem[{Liu et~al.(2020{\natexlab{b}})Liu, Yan, Gong, Qi, Zhang, Jiao, Chen,
  Fu, Shou, Gong, Wang, Chen, Jiang, Lv, Zhang, Wu, Zhou, and
  Duan}]{DBLP:journals/corr/abs-2011-11928}
Liu D, Yan Y, Gong Y, Qi W, Zhang H, Jiao J, Chen W, Fu J, Shou L, Gong M, Wang
  P, Chen J, Jiang D, Lv J, Zhang R, Wu W, Zhou M, Duan N (2020{\natexlab{b}})
  {GLGE:} {A} new general language generation evaluation benchmark. CoRR
  abs/2011.11928, \urlprefix\url{https://arxiv.org/abs/2011.11928},
  \eprint{2011.11928}

\bibitem[{Liu et~al.(2020{\natexlab{c}})Liu, Wang, Niu, Wu, Che, and
  Liu}]{DBLP:conf/acl/LiuWNWCL20}
Liu Z, Wang H, Niu Z, Wu H, Che W, Liu T (2020{\natexlab{c}}) Towards
  conversational recommendation over multi-type dialogs. In: Jurafsky D, Chai
  J, Schluter N, Tetreault JR (eds) Proceedings of the 58th Annual Meeting of
  the Association for Computational Linguistics, {ACL} 2020, Online, July 5-10,
  2020, Association for Computational Linguistics, pp 1036--1049,
  \doi{10.18653/v1/2020.acl-main.98},
  \urlprefix\url{https://doi.org/10.18653/v1/2020.acl-main.98}

\bibitem[{Loshchilov and Hutter(2017)}]{DBLP:journals/corr/abs-1711-05101}
Loshchilov I, Hutter F (2017) Fixing weight decay regularization in adam. CoRR
  abs/1711.05101, \urlprefix\url{http://arxiv.org/abs/1711.05101},
  \eprint{1711.05101}

\bibitem[{Mihaylov and Frank(2018)}]{DBLP:conf/acl/FrankM18}
Mihaylov T, Frank A (2018) Knowledgeable reader: Enhancing cloze-style reading
  comprehension with external commonsense knowledge. In: Gurevych I, Miyao Y
  (eds) Proceedings of the 56th Annual Meeting of the Association for
  Computational Linguistics, {ACL} 2018, Melbourne, Australia, July 15-20,
  2018, Volume 1: Long Papers, Association for Computational Linguistics, pp
  821--832, \doi{10.18653/v1/P18-1076},
  \urlprefix\url{https://www.aclweb.org/anthology/P18-1076/}

\bibitem[{Nogueira et~al.(2019)Nogueira, Yang, Cho, and
  Lin}]{DBLP:journals/corr/abs-1910-14424}
Nogueira R, Yang W, Cho K, Lin J (2019) Multi-stage document ranking with
  {BERT}. CoRR abs/1910.14424, \urlprefix\url{http://arxiv.org/abs/1910.14424},
  \eprint{1910.14424}

\bibitem[{Papineni et~al.(2002)Papineni, Roukos, Ward, and
  Zhu}]{DBLP:conf/acl/PapineniRWZ02}
Papineni K, Roukos S, Ward T, Zhu W (2002) Bleu: a method for automatic
  evaluation of machine translation. In: Proceedings of the 40th Annual Meeting
  of the Association for Computational Linguistics, July 6-12, 2002,
  Philadelphia, PA, {USA}, {ACL}, pp 311--318, \doi{10.3115/1073083.1073135},
  \urlprefix\url{https://www.aclweb.org/anthology/P02-1040/}

\bibitem[{Qi et~al.(2020)Qi, Yan, Gong, Liu, Duan, Chen, Zhang, and
  Zhou}]{DBLP:conf/emnlp/QiYGLDCZ020}
Qi W, Yan Y, Gong Y, Liu D, Duan N, Chen J, Zhang R, Zhou M (2020) Prophetnet:
  Predicting future n-gram for sequence-to-sequence pre-training. In: Cohn T,
  He Y, Liu Y (eds) Proceedings of the 2020 Conference on Empirical Methods in
  Natural Language Processing: Findings, {EMNLP} 2020, Online Event, 16-20
  November 2020, Association for Computational Linguistics, pp 2401--2410,
  \doi{10.18653/v1/2020.findings-emnlp.217},
  \urlprefix\url{https://doi.org/10.18653/v1/2020.findings-emnlp.217}

\bibitem[{Radford et~al.(2019)Radford, Wu, Child, Luan, Amodei, and
  Sutskever}]{radford2019language}
Radford A, Wu J, Child R, Luan D, Amodei D, Sutskever I (2019) Language models
  are unsupervised multitask learners. OpenAI blog 1(8):9

\bibitem[{Reimers and Gurevych(2019)}]{DBLP:conf/emnlp/ReimersG19}
Reimers N, Gurevych I (2019) Sentence-bert: Sentence embeddings using siamese
  bert-networks. In: Inui K, Jiang J, Ng V, Wan X (eds) Proceedings of the 2019
  Conference on Empirical Methods in Natural Language Processing and the 9th
  International Joint Conference on Natural Language Processing, {EMNLP-IJCNLP}
  2019, Hong Kong, China, November 3-7, 2019, Association for Computational
  Linguistics, pp 3980--3990, \doi{10.18653/v1/D19-1410},
  \urlprefix\url{https://doi.org/10.18653/v1/D19-1410}

\bibitem[{Vaswani et~al.(2017)Vaswani, Shazeer, Parmar, Uszkoreit, Jones,
  Gomez, Kaiser, and Polosukhin}]{DBLP:conf/nips/VaswaniSPUJGKP17}
Vaswani A, Shazeer N, Parmar N, Uszkoreit J, Jones L, Gomez AN, Kaiser L,
  Polosukhin I (2017) Attention is all you need. In: Guyon I, von Luxburg U,
  Bengio S, Wallach HM, Fergus R, Vishwanathan SVN, Garnett R (eds) Advances in
  Neural Information Processing Systems 30: Annual Conference on Neural
  Information Processing Systems 2017, December 4-9, 2017, Long Beach, CA,
  {USA}, pp 5998--6008,
  \urlprefix\url{http://papers.nips.cc/paper/7181-attention-is-all-you-need}

\bibitem[{Vinyals and Le(2015)}]{vinyals2015neural}
Vinyals O, Le Q (2015) A neural conversational model. arXiv preprint
  arXiv:150605869

\bibitem[{Wang et~al.(2015)Wang, Li, Lei, Huang, Yin, and
  Pong}]{wang2015constructing}
Wang F, Li X, Lei W, Huang C, Yin M, Pong TC (2015) Constructing learning maps
  for lecture videos by exploring wikipedia knowledge. In: Pacific Rim
  Conference on Multimedia, Springer, pp 559--569

\bibitem[{Wang et~al.(2020)Wang, Ke, Zheng, Huang, Jiang, Zhu, and
  Huang}]{DBLP:conf/nlpcc/WangKZHJZH20}
Wang Y, Ke P, Zheng Y, Huang K, Jiang Y, Zhu X, Huang M (2020) A large-scale
  chinese short-text conversation dataset. In: Zhu X, Zhang M, Hong Y, He R
  (eds) Natural Language Processing and Chinese Computing - 9th {CCF}
  International Conference, {NLPCC} 2020, Zhengzhou, China, October 14-18,
  2020, Proceedings, Part {I}, Springer, Lecture Notes in Computer Science, vol
  12430, pp 91--103, \doi{10.1007/978-3-030-60450-9\_8},
  \urlprefix\url{https://doi.org/10.1007/978-3-030-60450-9\_8}

\bibitem[{Zhang et~al.(2020{\natexlab{a}})Zhang, Liu, Lv, and
  Luo}]{DBLP:conf/acl/ZhangLLL20}
Zhang H, Liu D, Lv J, Luo C (2020{\natexlab{a}}) Let's be humorous: Knowledge
  enhanced humor generation. In: Rijhwani S, Liu J, Wang Y, Dror R (eds)
  Proceedings of the 58th Annual Meeting of the Association for Computational
  Linguistics: Student Research Workshop, {ACL} 2020, Online, July 5-10, 2020,
  Association for Computational Linguistics, pp 156--161,
  \urlprefix\url{https://www.aclweb.org/anthology/2020.acl-srw.21/}

\bibitem[{Zhang et~al.(2020{\natexlab{b}})Zhang, Sun, Galley, Chen, Brockett,
  Gao, Gao, Liu, and Dolan}]{DBLP:conf/acl/ZhangSGCBGGLD20}
Zhang Y, Sun S, Galley M, Chen Y, Brockett C, Gao X, Gao J, Liu J, Dolan B
  (2020{\natexlab{b}}) {DIALOGPT} : Large-scale generative pre-training for
  conversational response generation. In: {\c{C}}elikyilmaz A, Wen T (eds)
  Proceedings of the 58th Annual Meeting of the Association for Computational
  Linguistics: System Demonstrations, {ACL} 2020, Online, July 5-10, 2020,
  Association for Computational Linguistics, pp 270--278,
  \doi{10.18653/v1/2020.acl-demos.30},
  \urlprefix\url{https://doi.org/10.18653/v1/2020.acl-demos.30}

\bibitem[{Zhang et~al.(2021)Zhang, Zhang, Wang, Liang, Lei, Sun, Jatowt, and
  Yang}]{zhang2021generalized}
Zhang Y, Zhang X, Wang J, Liang H, Lei W, Sun Z, Jatowt A, Yang Z (2021)
  Generalized relation learning with semantic correlation awareness for link
  prediction. In: Proceedings of the AAAI Conference on Artificial Intelligence

\bibitem[{Zhang et~al.(2019)Zhang, Han, Liu, Jiang, Sun, and
  Liu}]{DBLP:conf/acl/ZhangHLJSL19}
Zhang Z, Han X, Liu Z, Jiang X, Sun M, Liu Q (2019) {ERNIE:} enhanced language
  representation with informative entities. In: Korhonen A, Traum DR,
  M{\`{a}}rquez L (eds) Proceedings of the 57th Conference of the Association
  for Computational Linguistics, {ACL} 2019, Florence, Italy, July 28- August
  2, 2019, Volume 1: Long Papers, Association for Computational Linguistics, pp
  1441--1451, \doi{10.18653/v1/p19-1139},
  \urlprefix\url{https://doi.org/10.18653/v1/p19-1139}

\bibitem[{Zheng et~al.(2020)Zheng, Cao, Jiang, and
  Huang}]{DBLP:conf/emnlp/ZhengCJH20}
Zheng C, Cao Y, Jiang D, Huang M (2020) Difference-aware knowledge selection
  for knowledge-grounded conversation generation. In: Cohn T, He Y, Liu Y (eds)
  Proceedings of the 2020 Conference on Empirical Methods in Natural Language
  Processing: Findings, {EMNLP} 2020, Online Event, 16-20 November 2020,
  Association for Computational Linguistics, pp 115--125,
  \doi{10.18653/v1/2020.findings-emnlp.11},
  \urlprefix\url{https://doi.org/10.18653/v1/2020.findings-emnlp.11}

\bibitem[{Zhou et~al.(2020)Zhou, Zheng, Huang, Huang, and
  Zhu}]{DBLP:conf/acl/ZhouZHHZ20}
Zhou H, Zheng C, Huang K, Huang M, Zhu X (2020) Kdconv: {A} chinese
  multi-domain dialogue dataset towards multi-turn knowledge-driven
  conversation. In: Jurafsky D, Chai J, Schluter N, Tetreault JR (eds)
  Proceedings of the 58th Annual Meeting of the Association for Computational
  Linguistics, {ACL} 2020, Online, July 5-10, 2020, Association for
  Computational Linguistics, pp 7098--7108,
  \doi{10.18653/v1/2020.acl-main.635},
  \urlprefix\url{https://doi.org/10.18653/v1/2020.acl-main.635}

\end{thebibliography}

\end{document}